%% file: emnlp2015.tex
\documentclass[11pt]{article}
\usepackage{acl2015}
\usepackage{times}
\usepackage{url}
\usepackage{latexsym}
\usepackage{amsmath,amsthm, amssymb, latexsym}
\usepackage{qtree}
\usepackage{xspace}
\usepackage[usenames,dvipsnames,svgnames,table]{xcolor}
\usepackage[breaklinks=true,letterpaper=true,colorlinks,bookmarks=false,citecolor=MidnightBlue,linkcolor=BrickRed]{hyperref}

\usepackage{tikz}
\usetikzlibrary{calc}

\usepackage{color}
\usepackage{multirow}
\usepackage{array}
\usepackage{graphicx}
\usepackage{caption}
\usepackage{subcaption}
\usepackage{tabularx}

\usepackage{bigstrut}
\graphicspath{{pics/}}
\DeclareGraphicsExtensions{.jpg,.jpeg,.png,.pdf}

\definecolor{Red}{rgb}{1,0,0}
\definecolor{Green}{rgb}{0,0.69,0}
\definecolor{Blue}{rgb}{0,0,1}
\definecolor{LightBlue}{rgb}{0,0.5,1}
\definecolor{lightBlue}{rgb}{0,0.5,1}
\definecolor{Skin}{rgb}{1,0.71,0.69}
\definecolor{Grey}{rgb}{0.5,0.5,0.5}
\definecolor{LightGrey}{rgb}{0.6,0.6,0.6}
\definecolor{Black}{rgb}{0,0,0}
\definecolor{White}{rgb}{1,1,1}
\newcommand{\red}{\color{Red}}

\newcommand{\lightBlue}{\color{LightBlue}}

\newcommand{\grey}{\color{Grey}}

\newcommand{\ambiguityExample}[9]{%
  & \hspace{3em} #2 & \hspace{3em} #3 \\[1ex]
  & \multicolumn{2}{c}{\begin{tabular}{@{}c@{\hspace{0.2em}}c@{\hspace{0.2em}}c@{\hspace{1.2em}}c@{\hspace{0.2em}}c@{\hspace{0.2em}}c@{}}
        \includegraphics[width=0.15\textwidth]{#4}& \includegraphics[width=0.15\textwidth]{#5}& \includegraphics[width=0.15\textwidth]{#6}&
        \includegraphics[width=0.15\textwidth]{#7}& \includegraphics[width=0.15\textwidth]{#8}& \includegraphics[width=0.15\textwidth]{#9}
  \end{tabular}}\\[7ex]
}

\DeclareUrlCommand\emailurl{}
\newcommand{\email}[1]{\href{mailto:#1}{\emailurl{#1}}}

\def\etal{et al.\@\xspace}
\def\eg{e.g.\@\xspace}
\def\ie{i.e.\@\xspace}

\title{Do You See What I Mean?\\Visual Resolution of Linguistic Ambiguities}

\author{Yevgeni Berzak \\
  CSAIL MIT \\
  \email{berzak@mit.edu} \\\And
 Andrei Barbu \\
 CSAIL MIT \\
 \email{andrei@0xab.com} \\\And
  Daniel Harari \\ 
  CSAIL MIT \\
  \email{hararid@mit.edu} \\\AND
  Boris Katz \\
  CSAIL MIT \\
  \email{boris@mit.edu} \\\And
  Shimon Ullman \\
  Weizmann Institute of Science \\
  \email{shimon.ullman@weizmann.ac.il} \\ }

\date{}

\begin{document}
\maketitle
\begin{abstract}
Understanding language goes hand in hand with the ability to integrate complex 
contextual information obtained via perception. In this work, we present a novel 
task for grounded language understanding: disambiguating a sentence 
given a visual scene which depicts one of the possible interpretations of that 
sentence. To this end, we introduce a new multimodal corpus containing ambiguous 
sentences, representing a wide range of syntactic, semantic and discourse ambiguities,
coupled with videos that visualize the different interpretations for each sentence. 
We address this task by extending a vision model which determines if a
sentence is depicted by a video. We demonstrate how such a model
can be adjusted to recognize different interpretations of the same underlying sentence,
allowing to disambiguate sentences in a unified fashion across the different ambiguity types.

\end{abstract}

\section{Introduction}
\label{sec:intro}

Ambiguity is one of the defining characteristics of human languages, and language 
understanding crucially relies on the ability to obtain unambiguous representations 
of linguistic content. While some ambiguities can be resolved using intra-linguistic
contextual cues, the disambiguation of many linguistic constructions 
requires integration of world knowledge and perceptual information obtained from 
other modalities.

In this work, we focus on the problem of grounding language in the visual modality, 
and introduce a novel task for language understanding which requires resolving 
linguistic ambiguities by utilizing the visual context in which the linguistic content 
is expressed. This type of inference is frequently called for in human communication
that occurs in a visual environment, and is crucial for language acquisition, 
when much of the linguistic content refers to the visual surroundings of the child \cite{snow1972mothers}.
 
Our task is also fundamental to the problem of grounding vision in language, 
by focusing on phenomena of linguistic ambiguity, which are prevalent in language, 
but typically overlooked when using language as a medium for expressing understanding of visual content.
Due to such ambiguities, a superficially appropriate description of a visual scene 
may in fact not be sufficient for demonstrating a correct understanding of the relevant visual content.
Our task addresses this issue by introducing a deep validation protocol for visual understanding, requiring 
not only providing a surface description of a visual activity but also 
demonstrating structural understanding at the levels of syntax, semantics and discourse.

To enable the systematic study of visually grounded processing of ambiguous language, 
we create a new corpus, 
\emph{LAVA} (Language and Vision Ambiguities).
This corpus contains sentences with linguistic ambiguities that can only be resolved
using external information. The sentences are paired with short videos that visualize different 
interpretations of each sentence. Our sentences encompass a wide range of syntactic, semantic
and discourse ambiguities, including ambiguous prepositional and verb phrase attachments, 
conjunctions, logical forms, anaphora and ellipsis. Overall, the corpus contains 
237 sentences, with 2 to 3 interpretations per sentence, and an average of 3.37 videos that depict 
visual variations of each sentence interpretation, corresponding to a total of
1679 videos.

Using this corpus, we address the problem of selecting the
interpretation of an ambiguous sentence that matches the
content of a given video. Our approach for tackling this task extends
the \emph{sentence tracker} introduced in \cite{siddharth2014}. The
sentence tracker produces a score which determines if a sentence is
depicted by a video. This earlier work had no concept of ambiguities;
it assumed that every sentence had a single
interpretation. We extend this approach to represent multiple interpretations 
of a sentence, enabling us to pick the interpretation that is most compatible
with the video.

To summarize, the contributions of this paper are threefold. First, we introduce 
a new task for visually grounded language understanding, in which an 
ambiguous sentence has to be disambiguated using a visual depiction of the sentence's 
content. Second, we release a multimodal corpus of sentences coupled with videos 
which covers a wide range of linguistic ambiguities, and enables a systematic study
of linguistic ambiguities in visual contexts. Finally, we present a computational
model which disambiguates the sentences in our corpus with an accuracy of 75.36\%.

\section{Related Work}
\label{sec:related_work}

Previous language and vision studies focused on the development of  
multimodal word and sentence representations \cite{bruni2012,socher2013,lapata2014,gong2014,lazaridou15},
as well as methods for describing images and videos in natural language 
\cite{farhadi2010,kulkarni2011,mitchell2012,socher2014,thomason2014,karpathy2014,siddharth2014,venugopalan2015,vinyals2015}.
While these studies handle important challenges in multimodal processing of language and vision, 
they do not provide explicit modeling of linguistic ambiguities.

Previous work relating ambiguity in language to the visual modality addressed the problem of
word sense disambiguation \cite{barnard2003}. However, this work is limited to context independent interpretation of individual words, 
and does not consider structure-related ambiguities.
Discourse ambiguities were previously studied in work on multimodal coreference resolution \cite{ramanathan2014,kong2014}.
Our work expands this line of research, and addresses further discourse ambiguities in the interpretation 
of ellipsis. More importantly, to the best of our knowledge our study is the first to present a systematic treatment of syntactic and 
semantic sentence level ambiguities in the context of language and vision. 

The interactions between linguistic and visual information in human sentence 
processing have been extensively studied in psycholinguistics and cognitive psychology \cite{tanenhaus1995}.
A considerable fraction of this work focused on the processing of ambiguous language \cite{spivey2002,coco2015},
providing evidence for the importance of visual information for linguistic ambiguity resolution by humans.
Such information is also vital during language acquisition, when much of the linguistic content perceived
by the child refers to their immediate visual environment \cite{snow1972mothers}. Over time, children develop
mechanisms for grounded disambiguation of language, manifested among others by the usage of iconic gestures 
when communicating ambiguous linguistic content \cite{kidd2009children}.
Our study leverages such insights to develop a complementary framework that enables addressing the challenge
of visually grounded disambiguation of language in the realm of artificial intelligence.

\section{Task}
\label{sec:task}
In this work we provide a concrete framework for the study of language 
understanding with visual context by introducing the task of grounded language disambiguation. 
This task requires to choose the correct linguistic representation of a sentence given a visual 
context depicted in a video. Specifically, provided with a sentence, $n$ candidate 
interpretations of that sentence and a video that depicts the content 
of the sentence, one needs to choose the interpretation that corresponds to the content of 
the video. 

\begin{figure}
        \centering
        \begin{subfigure}[b]{0.23\textwidth}
          \scalebox{0.4}{\Tree [.S [.NP [.NNP Sam ] ] [.VP [.VP [.V [.VBD approached ] ]  [.NP [.DT the ] [.NN chair ] ] ] [.PP [.IN with ] [.NP [.DT a ] [.NN bag ] ] ] ] ]}
          \caption{First interpretation}
          \label{fig:example:tree1}
        \end{subfigure}%
        \begin{subfigure}[b]{0.23\textwidth}
          \scalebox{0.4}{\Tree [.S [.NP [.NNP Sam ] ] [.VP [.V [.VBD approached ] ] [.NP  [.NP [.DT the ] [.NN chair ] ]  [.PP [.IN with ] [.NP [.DT a ] [.NN bag ] ] ] ] ] ]}
          \caption{Second interpretation}
          \label{fig:example:tree2}
        \end{subfigure}\\[2ex]
        \begin{subfigure}[b]{0.48\textwidth}
          \begin{tabular}{@{}c@{\hspace{0.3em}}c@{\hspace{0.3em}}c@{}}
            \includegraphics[width=0.32\textwidth]{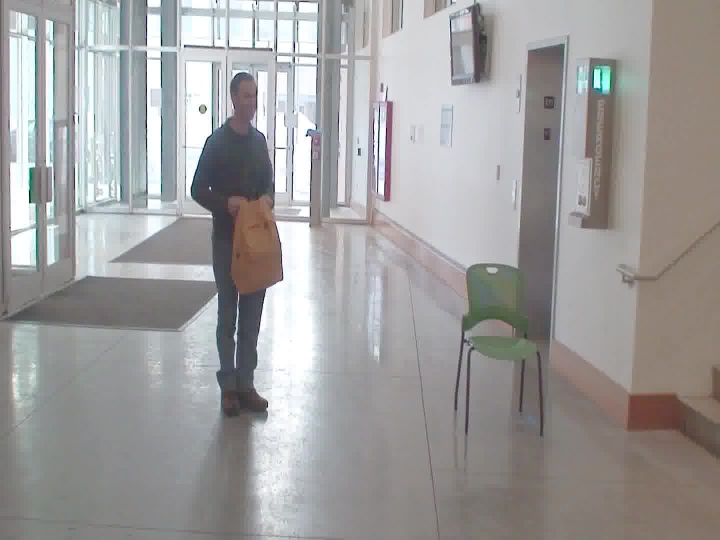}&
            \includegraphics[width=0.32\textwidth]{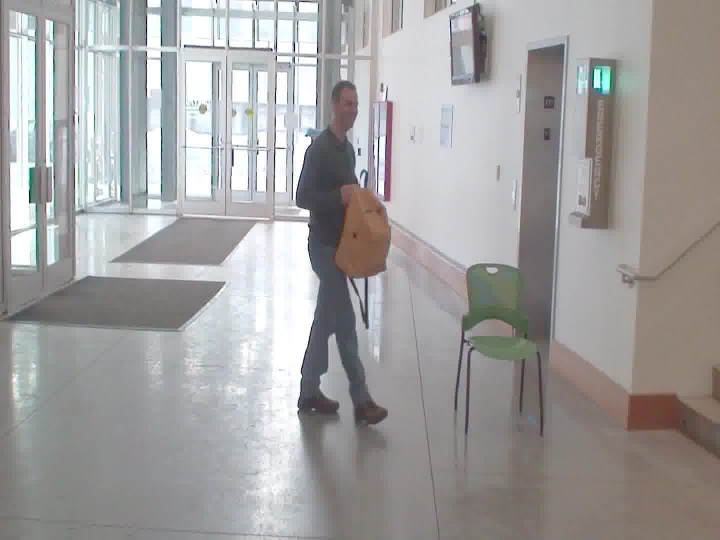}&
            \includegraphics[width=0.32\textwidth]{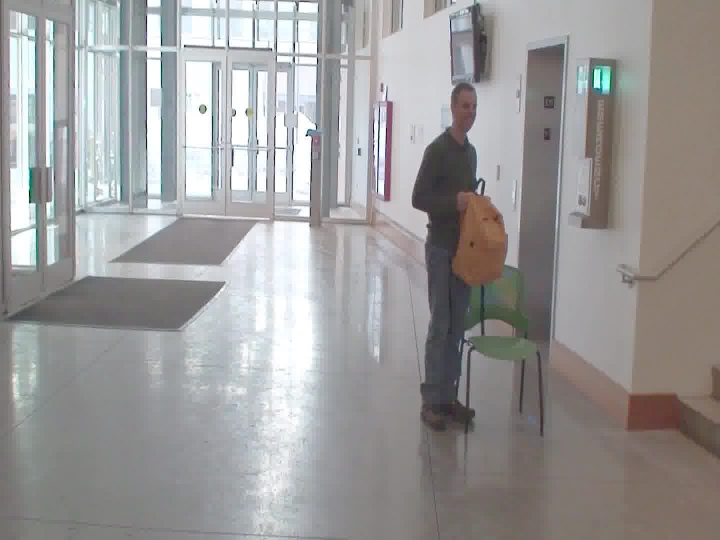}
          \end{tabular}
          \caption{Visual context}
          \label{fig:example:vis1}
        \end{subfigure}
        \caption{An example of the visually grounded language disambiguation task.
          Given the sentence ``Sam approached the chair with a bag'', two potential parses, (a) and (b), correspond to two different semantic interpretations.
	  In the first interpretation Sam has the bag, while in the second reading the bag is on the chair.
          The task is to select the correct interpretation given the visual context (c).}\label{fig:task}
\end{figure}

To illustrate this task, consider the example in figure~\ref{fig:task}, where we are given
the sentence ``Sam approached the chair with a bag'' along with two different linguistic interpretations.
In the first interpretation, which corresponds to parse~\ref{fig:task}(a), Sam has the bag. In the second
interpretation associated with parse~\ref{fig:task}(b), the bag is on the chair rather than with Sam. 
Given the visual context from figure~\ref{fig:task}(c), the task is to choose which interpretation is most appropriate 
for the sentence.

\section{Approach Overview}
\label{sec:overview}

To address the grounded language disambiguation task, we use a compositional
approach for determining if a specific interpretation of a sentence is
depicted by a video.
In this framework, described in detail in section \ref{sec:model}, a
sentence and an accompanying interpretation encoded in first order logic, 
give rise to a grounded model that matches a video against the provided 
sentence interpretation.

The model is comprised of Hidden Markov Models (HMMs) which encode
the semantics of words, and trackers which locate objects in video
frames.
To represent an interpretation of a sentence, word models are combined
with trackers through a cross-product which respects the semantic representation 
of the sentence to create a single model which recognizes that interpretation.

Given a sentence, we construct an HMM based representation for each
interpretation of that sentence.
We then detect candidate locations for objects in every frame of the
video.
Together the reforestation for the sentence and the candidate object
locations are combined to form a model which can determine if a given
interpretation is depicted by the video.
We test each interpretation and report the interpretation with highest
likelihood.

\section{Corpus}
\label{sec:corpus}

\begin{figure}
   \centering
   \begin{subfigure}[b]{0.23\textwidth}
     \scalebox{0.4}{\Tree [.S  [.NP [.NNP Bill ] ] [.VP [.VBD held ] [.NP [.DT the ] [.NP [.JJ green ] [.NP [.NN chair ] [.CC and ] [.NN bag ] ] ] ] ] ]} \\
     \caption{First interpretation}
     \label{fig:example:tree1}
   \end{subfigure}%
   \begin{subfigure}[b]{0.23\textwidth}
     \scalebox{0.4}{\Tree [.S  [.NP [.NNP Bill ] ] [.VP [.VBD held ] [.NP [.DT the ] [.NP [.NP [.JJ green ] [.NN chair ] ]  [.CC and ] [.NN bag ] ] ] ] ] }
     \caption{Second interpretation}
     \label{fig:example:tree2}
   \end{subfigure}\\[2ex]
   \begin{subfigure}[b]{0.23\textwidth}
     \includegraphics[width=0.9\textwidth]{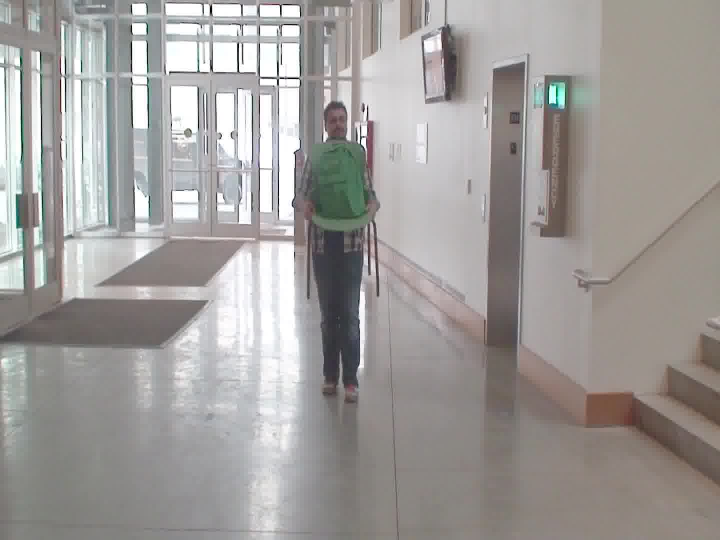}\\
     \caption{First visual context}
     \label{fig:example:vis1}
   \end{subfigure}
   \begin{subfigure}[b]{0.23\textwidth}
     \includegraphics[width=0.9\textwidth]{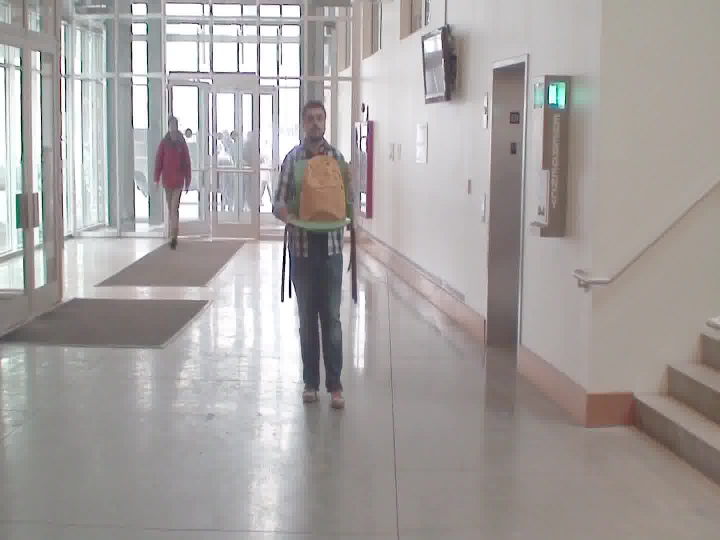}\\
     \caption{Second visual context}
     \label{fig:example:vis2}
   \end{subfigure}
   \caption{Linguistic and visual interpretations of the sentence ``Bill held the green chair and bag''.
	    In the first interpretation (a,c) both the chair and bag are green, while in the second interpretation (b,d) 
	   only the chair is green and the bag has a different color.}\label{fig:example}
\end{figure}

To enable a systematic study of linguistic ambiguities that are grounded in vision, 
we compiled a corpus with ambiguous sentences describing visual actions. 
The sentences are formulated such that the correct linguistic interpretation
of each sentence can only be determined using external, non-linguistic, information about the 
depicted activity. For example, in the sentence ``Bill held the green chair 
and bag'', the correct scope of ``green'' can only be determined by integrating 
additional information about the color of the bag. This information is provided 
in the accompanying videos, which visualize the possible interpretations of 
each sentence. Figure~\ref{fig:example} presents the syntactic parses 
for this example along with frames from the respective videos. Although our videos 
contain visual uncertainty, they are not ambiguous with respect to the linguistic 
interpretation they are presenting, and hence a video always corresponds to a 
single candidate representation of a sentence.

The corpus covers a wide range of well known syntactic, semantic and discourse 
ambiguity classes. While the ambiguities are associated with various types, 
different sentence interpretations always represent distinct sentence meanings, 
and are hence encoded semantically using first order logic. 
For syntactic and discourse ambiguities we also provide 
an additional, ambiguity type specific encoding as described below.

\begin{itemize}
\item \textbf{Syntax} Syntactic ambiguities include Prepositional Phrase (PP) attachments,
Verb Phrase (VP) attachments, and ambiguities in the interpretation of conjunctions.
In addition to logical forms, sentences with syntactic ambiguities are also accompanied with 
Context Free Grammar (CFG) parses of the candidate interpretations, generated from a 
deterministic CFG parser. 
\item \textbf{Semantics} The corpus addresses several classes of semantic quantification 
ambiguities, in which a syntactically unambiguous sentence may correspond to different logical forms. For each such
sentence we provide the respective logical forms.
\item \textbf{Discourse} The corpus contains two types of discourse ambiguities, Pronoun Anaphora
and Ellipsis, offering examples comprising two sentences. 
In \textbf{anaphora} ambiguity cases, an ambiguous pronoun in the second sentence 
is given its candidate antecedents in the first sentence, as well as a corresponding 
logical form for the meaning of the second sentence.
In \textbf{ellipsis} cases, a part of the second sentence, which can constitute either 
the subject and the verb, or the verb and the object, is omitted. We provide both 
interpretations of the omission in the form of a single unambiguous sentence, and its 
logical form, which combines the meanings of the first and the second sentences.
\end{itemize}

Table~\ref{tab:sentence-examples} lists examples of the different ambiguity classes, 
along with the candidate interpretations of each example. 

\begin{table}
\scriptsize
\begin{center}
\begin{tabular}{llll}
                                                    &\bf Ambiguity & \bf Templates & \bf \# \\
\cline{2-4}\bigstrut
\multirow{4}{*}{{\rotatebox[origin=c]{90}{Syntax}}} & PP & NNP V DT [JJ] $\text{NN}_{1}$ IN DT [JJ] $\text{NN}_{2}$. &48\\ 
\cline{2-4}\bigstrut
                                                    & VP & $\text{NNP}_{1}$ V [IN] $\text{NNP}_{2}$ V [JJ] NN. &60\\
\cline{2-4}\bigstrut
                                                    & Conjunction   & $\text{NNP}_{1}$ [and $\text{NNP}_{2}$] V DT JJ $\text{NN}_{1}$ and $\text{NN}_{2}$. &40\\
	                                            &		    & NNP V DT $\text{NN}_{1}$ or DT $\text{NN}_{2}$ and DT $\text{NN}_{3}$. &\\
\hline\bigstrut
\multirow{4}{*}{{\rotatebox[origin=c]{90}{Semantics}}}& Logical Form & $\text{NNP}_{1}$ and $\text{NNP}_{2}$ V a NN.       &35\\
                                                      &	 & Someone V the NNS.	        &\\[5ex]
\hline\bigstrut
\multirow{4}{*}{{\rotatebox[origin=c]{90}{Discourse}}}& Anaphora & NNP V DT $\text{NN}_{1}$ and DT $\text{NN}_{2}$. It is JJ. &36\\
\cline{2-4}\bigstrut
                                                      & Ellipsis & $\text{NNP}_{1}$ V $\text{NNP}_{2}$. Also $\text{NNP}_{3}$.               &18\\[5ex]
\end{tabular}
\end{center}
\caption{POS templates for generating the sentences in our corpus. The rightmost column represents the number of sentences in each category.
The sentences are produced by replacing the POS tags with all the visually applicable assignments of lexical items from the corpus lexicon shown in table~\ref{tab:lexicon}.}\label{tab:templates}
\end{table} 

\begin{table*}[ht]
\scriptsize
\begin{center}
\begin{tabularx}{\textwidth}{ll>{\hsize=.7\hsize}X>{\hsize=1.3\hsize}Xl}
                                                    &\bf Ambiguity & \bf Example & \bf Linguistic interpretations & \bf Visual setups \\ 
\cline{1-5}\bigstrut

\multirow{13}{*}{{\rotatebox[origin=c]{90}{Syntax}}} & PP           & Claire left the green chair with a yellow bag.    & Claire [left the green chair] [with a yellow bag]. &The bag is with Claire.\\
						    &               &                             & Claire left [the green chair with a yellow bag]. & Bag is on the chair. \\
\cline{2-5}\bigstrut
                                                    & VP 	    & Claire looked at Bill picking up a chair.  & Claire looked at [Bill [picking up a chair]]. & Bill picks up the chair. \\
                                                    & 		    &                        & Claire [looked at Bill] [picking up a chair]. & Claire picks up the chair. \\
\cline{2-5}\bigstrut
                                                    & Conjunction   & Claire held a green bag and chair.    & Claire held a [green [bag and chair]]. & The chair is green.  \\
						    &		    &        & Claire held a [[green bag] and [chair]]. & The chair is not green. \\
	                                            &		    & Claire held the chair or the bag and the telescope.  & Claire held [[the chair] or [the bag and the telescope]]. & Claire holds the chair.\\
						    &		    & & Claire held [[the chair or the bag] and [the telescope]].& Claire holds the chair and the telescope.\\
\hline\bigstrut
\multirow{7}{*}{{\rotatebox[origin=c]{90}{Semantics}}}& Logical Form & Claire and Bill moved a chair. & $\textbf{chair}(x), \textbf{move}(\text{Claire}, x), \textbf{move}(\text{Bill}, x)$ & Claire and Bill move the same chair.\\
						      &	        &         		      & $\textbf{chair}(x), \textbf{chair}(y), \textbf{move}(\text{Claire}, x),$\newline$\textbf{move}(\text{Bill}, y), x\neq y$ & Claire and Bill move different chairs.\\ 
		                                      &         &  Someone moved the two chairs. & $\textbf{chair}(x), \textbf{chair}(y), x\neq y, \textbf{person}(u),$\newline\hspace*{1em}$\textbf{move}(u,x), \textbf{move}(u,y)$ & One person moves both chairs.\\
		                                      &         &                             &  $\textbf{chair}(x), \textbf{chair}(y), x\neq y, \textbf{person}(u), \textbf{person}(v),$\newline\hspace*{1em}$u \neq v,\textbf{move}(u,x), \textbf{move}(v,y)$ & Each chair moved by a different person.\\
\hline\bigstrut
\multirow{5
}{*}{{\rotatebox[origin=c]{90}{Discourse}}}& Anaphora & Claire held the bag and the  & It = bag   & The bag is yellow. \\
                                                      &         & chair. It is yellow.       & It = chair &  The chair is yellow. \\
\cline{2-5}\bigstrut
                                                      & Ellipsis & Claire looked at Bill.    & Claire looked at Bill and Sam.    & Claire looks at Bill and Sam. \\
                                                      &         & Also Sam.                  & Claire and Sam looked at Bill.   & Claire and Sam look at Bill. \\
\end{tabularx}
\end{center}
\caption{An overview of the different ambiguity types, along with examples of ambiguous sentences with their linguistic and visual interpretations. 
Note that similarly to semantic ambiguities, syntactic and discourse ambiguities are also provided with first order logic formulas for the resulting sentence interpretations. Table~\ref{tab:vision-examples} shows additional examples for each ambiguity type, with frames from sample videos corresponding to the different interpretations of each sentence.}\label{tab:sentence-examples}
\end{table*} 

The corpus is generated using Part of Speech (POS) tag sequence templates. For each template, 
the POS tags are replaced with lexical items from the corpus lexicon, 
described in table~\ref{tab:lexicon}, using all the visually applicable assignments. 
This generation process yields an overall of 237 sentences, of which 213 sentences
have 2 candidate interpretations, and 24 sentences have 3 interpretations.
%
%
Table~\ref{tab:templates} presents 
the corpus templates for each ambiguity class, along with the number of sentences 
generated from each template.

\begin{table*}[ht]
\small
\begin{center}
\begin{tabular}{lll}
\bf Syntactic Category & \bf Visual Category & \bf Words \\ \hline\bigstrut
    Nouns & Objects, People & chair, bag, telescope, someone, proper names \\ \hline\bigstrut
    Verbs & Actions & pick up, put down, hold, move (transitive), look at, approach, leave \\ \hline\bigstrut
    Prepositions & Spacial Relations & with, left of, right of, on \\ \hline\bigstrut
    Adjectives & Visual Properties & yellow, green \\
\end{tabular}
\end{center}
\caption{The lexicon used to instantiate the templates in figure~\ref{tab:templates} in order to generate the corpus.}\label{tab:lexicon}
\end{table*} 

\begin{table*}
\centering
\begin{tabularx}{\textwidth}{@{}lll@{}}
  \ambiguityExample{10}{PP Attachment}{Sam looked at Bill with a telescope.}{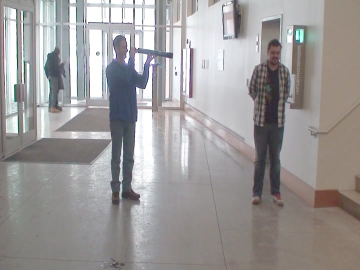}{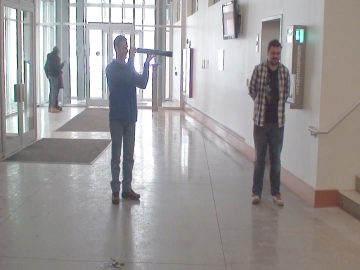}{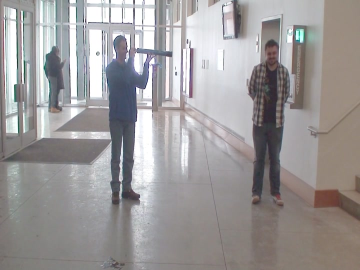}{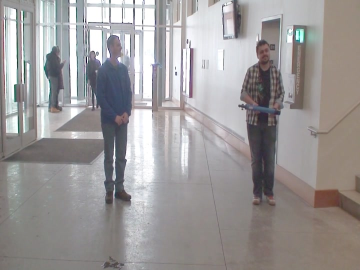}{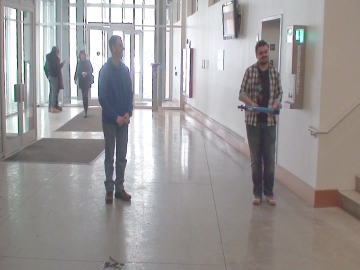}{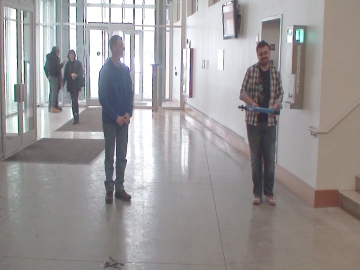}
  \ambiguityExample{12}{VP Attachment}{Bill approached the person holding a green chair.}{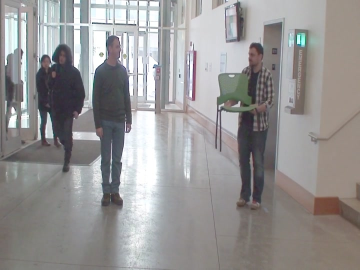}{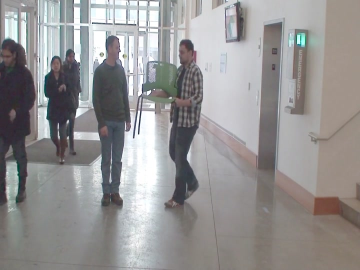}{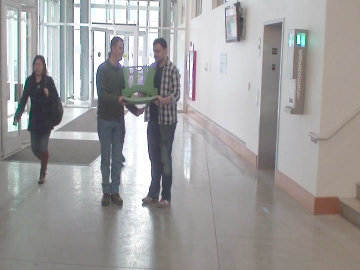}{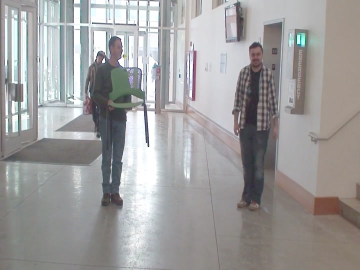}{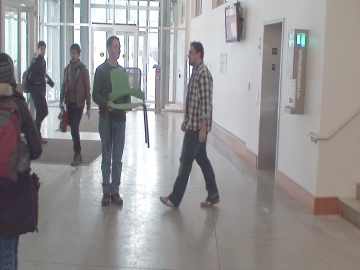}{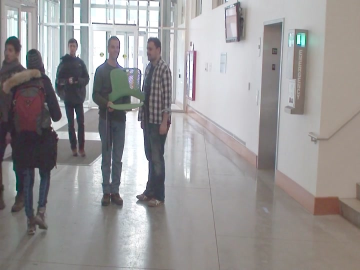}
  \ambiguityExample{12}{Conjunction}{Sam and Bill picked up the yellow bag and chair.}{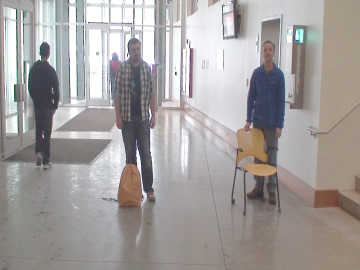}{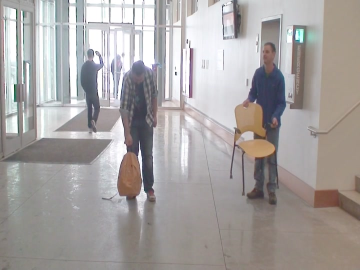}{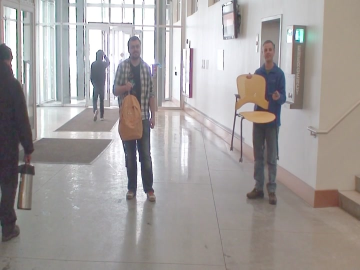}{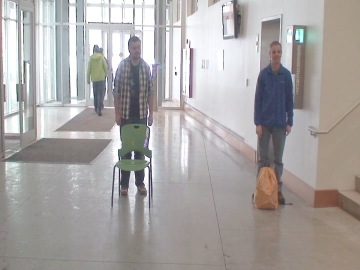}{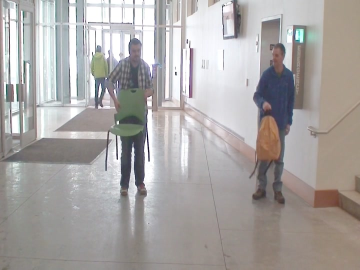}{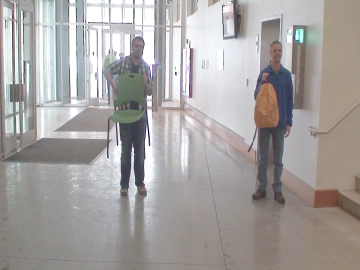}
  \ambiguityExample{12}{Logical Form}{Someone put down the bags.}{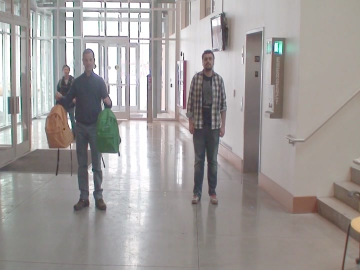}{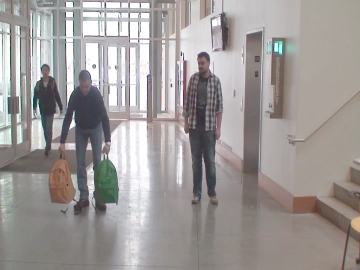}{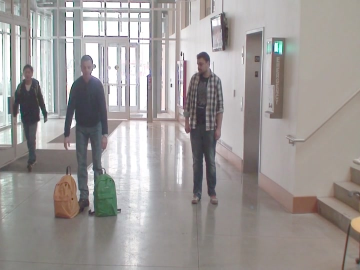}{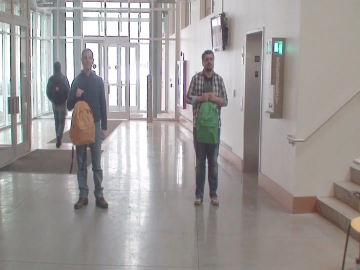}{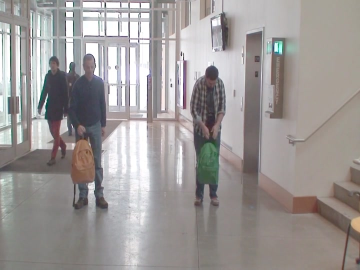}{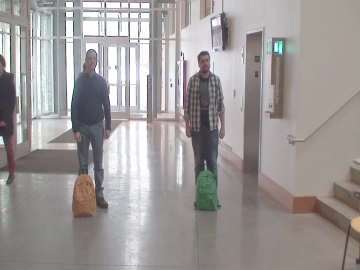}
  \ambiguityExample{12}{Anaphora}{Sam picked up the bag and the chair. It is yellow.}{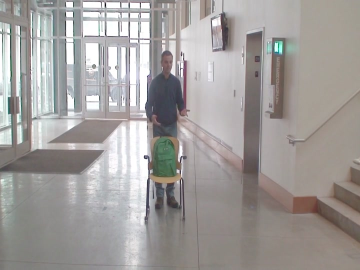}{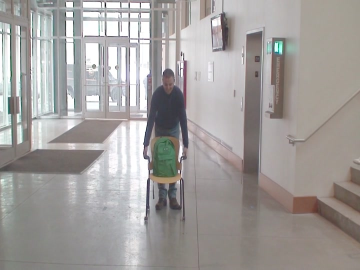}{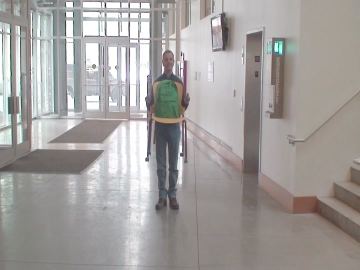}{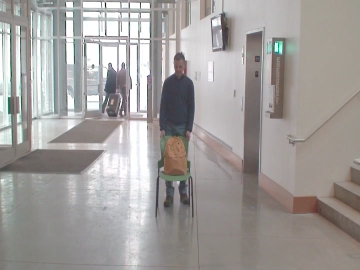}{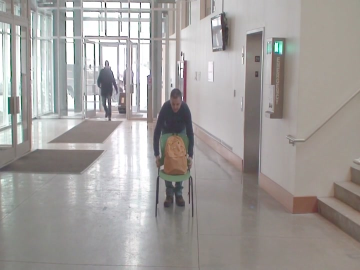}{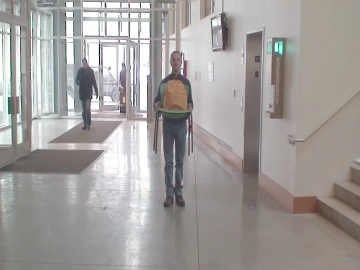}
  \ambiguityExample{12}{Ellipsis}{Sam left Bill. Also Clark.}{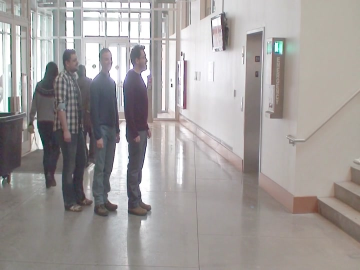}{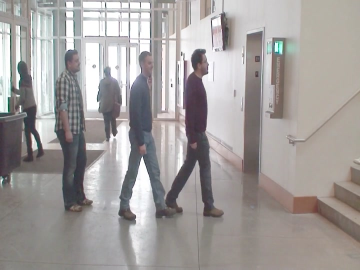}{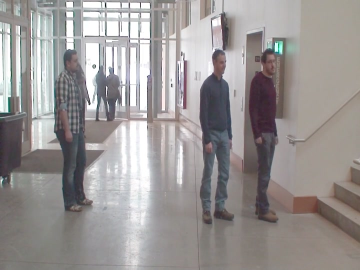}{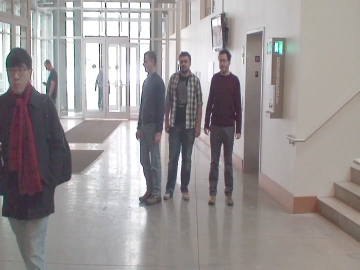}{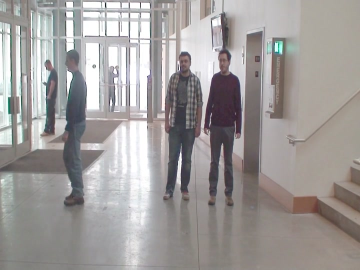}{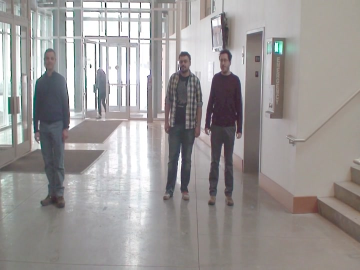}
\end{tabularx}
\caption{Examples of the six ambiguity classes described in table~\ref{tab:sentence-examples}. The example sentences have at least two interpretations, which
are depicted by different videos. Three frames from each such video are shown on the left and on the right below each sentence.}
\label{tab:vision-examples}
\end{table*}

The corpus videos are filmed in an indoor environment containing 
background objects and pedestrians.
To account for the manner of performing actions, videos are shot twice with different actors.
Whenever applicable, we also filmed the actions from two different directions 
(\eg approach from the left, and approach from the right). Finally, all videos were 
shot with two cameras from two different view points. Taking these variations into 
account, the resulting video corpus contains 7.1 videos per sentence and 3.37 videos 
per sentence interpretation, corresponding to a total of 1679 videos. The average video 
length is 3.02 seconds (90.78 frames), with in an overall of 1.4 hours of footage (152434 frames).

A custom corpus is required for this task because no existing corpus,
containing either videos or images, systematically covers multimodal
ambiguities.
Datasets such as \emph{UCF Sports} \cite{Rodriguez2008},
\emph{YouTube} \cite{liu2009recognizing}, and \emph{HMDB}
\cite{kuehne2011hmdb} which come out of the activity recognition community
are accompanied by action labels, not sentences, and do not
control for the content of the videos aside from the principal action
being performed.
Datasets for image and video captioning, such as \emph{MSCOCO}
\cite{lin2014microsoft} and \emph{TACOS} \cite{regneri2013grounding}, aim to
control for more aspects of the videos than just the main action being
performed but they do not provide the range of ambiguities discussed
here.
The closest dataset is that of Siddharth \etal (2014)
as it controls for object appearance, color, action, and direction of
motion, making it more likely to be suitable for evaluating
disambiguation tasks.
Unfortunately, that dataset was designed to avoid ambiguities, and therefore
is not suitable for evaluating the work described here.

\section{Model}
\label{sec:model}

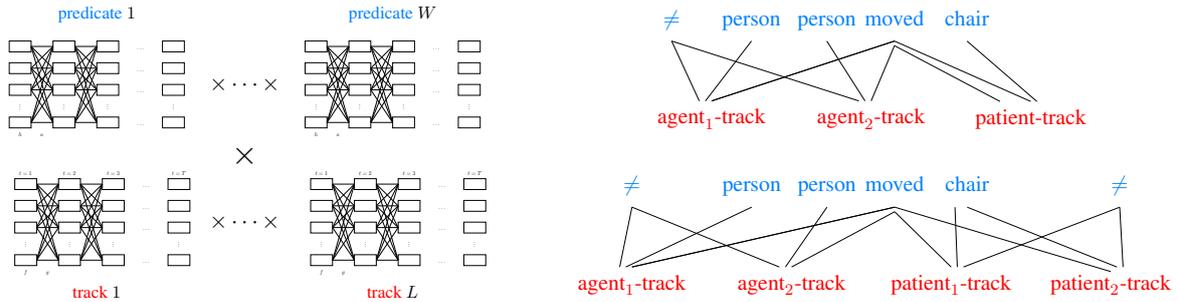
\begin{figure*}
  \centering
  \begin{tabular}{@{}c@{\hspace{6ex}}c@{}}
    \scalebox{0.6}{
      \begin{tabular}{@{}ccc@{}}
        {\lightBlue predicate} $1$&&{\lightBlue predicate} $W$\\[2ex]
        \scalebox{0.3}{\input{viterbi-event}}&
        \raisebox{30pt}{{\Large $\times\cdots\times$}}&
        \scalebox{0.3}{\input{viterbi-event}}\\
        &{\huge $\times$}&\\
        \scalebox{0.3}{\input{viterbi-object}}&
        \raisebox{30pt}{{\Large $\times\cdots\times$}}&
        \scalebox{0.3}{\input{viterbi-object}}\\
        {\red track} $1$&&{\red track} $L$\\
      \end{tabular}}
    &
    \begin{minipage}{0.5\textwidth}
      \centering
    \scalebox{0.75}{
    \begin{tikzpicture}[scale=7]
      \node[] (Claire) at (0.5,0.7) {\lightBlue person\strut};
      \node[] (and) at ($(Claire.east)+(0.1em,0)$) {\grey \strut};
      \node[] (Bill) at ($(and.east)+(0.1em,0)$) {\lightBlue person\strut};
      \node[] (moved) at ($(Bill.east)+(0.2em,0)$) {\lightBlue moved\strut};
      \node[] (a) at ($(moved.east)+(0.04em,0)$) {\grey \strut};
      \node[] (chair) at ($(a.east)+(0.14em,0)$) {\lightBlue chair\strut};
      \node[] (agent1) at (0.4, 0.45) {{\red $\text{agent}_1$-track}};
      \node[] (agent2) at (0.8, 0.45) {{\red $\text{agent}_2$-track}};
      \node[] (patient) at (1.2, 0.45) {{\red patient-track}};
      \node[] (neqA) at (0.3,0.7) {\lightBlue $\neq$\strut};
      \draw[-] ($(neqA.south)$) -- ($(agent1.north)-(0.08em,0)$);
      \draw[-] ($(neqA.south)$) -- ($(agent2.north)-(0.08em,0)$);
      \draw[-] ($(Claire.south)$) -- ($(agent1.north)+(-0.04em,0)$);
      \draw[-] ($(Bill.south)$) -- ($(agent2.north)+(-0.04em,0)$);
      \draw[-] ($(chair.south)$) -- ($(patient.north)+(0.04em,0)$);
      \draw[-] ($(moved.south)$) -- ($(agent1.north)$);
      \draw[-] ($(moved.south)$) -- ($(patient.north)$);
      \draw[-] ($(moved.south)$) -- ($(agent1.north)$);
      \draw[-] ($(moved.south)-(0,0.03em)$) -- ($(patient.north)-(0.2em,0)$);
      \draw[-] ($(moved.south)-(0,0.03em)$) -- ($(agent2.north)$);
    \end{tikzpicture}}\\[2ex]
  \scalebox{0.75}{
    \begin{tikzpicture}[scale=7]
      \node[] (Claire) at (0.7,0.7) {\lightBlue person\strut};
      \node[] (and) at ($(Claire.east)+(0.1em,0)$) {\grey \strut};
      \node[] (Bill) at ($(and.east)+(0.1em,0)$) {\lightBlue person\strut};
      \node[] (moved) at ($(Bill.east)+(0.2em,0)$) {\lightBlue moved\strut};
      \node[] (a) at ($(moved.east)+(0.04em,0)$) {\grey \strut};
      \node[] (chair) at ($(a.east)+(0.14em,0)$) {\lightBlue chair\strut};
      \node[] (agent1) at (0.4, 0.45) {{\red $\text{agent}_1$-track}};
      \node[] (agent2) at (0.8, 0.45) {{\red $\text{agent}_2$-track}};
      \node[] (patient1) at (1.2, 0.45) {{\red $\text{patient}_1$-track}};
      \node[] (patient2) at (1.6, 0.45) {{\red $\text{patient}_2$-track}};
      \node[] (neqA) at (0.4,0.7) {\lightBlue $\neq$\strut};
      \node[] (neq) at ($(chair.east)+(0.8em,0)$) {\lightBlue $\neq$\strut};
      \draw[-] ($(neqA.south)$) -- ($(agent1.north)-(0.08em,0)$);
      \draw[-] ($(neqA.south)$) -- ($(agent2.north)-(0.08em,0)$);
      \draw[-] ($(Claire.south)$) -- ($(agent1.north)+(-0.04em,0)$);
      \draw[-] ($(Bill.south)$) -- ($(agent2.north)+(-0.04em,0)$);
      \draw[-] ($(chair.south)-(0.08em,0)$) -- ($(patient1.north)+(0.04em,0)$);
      \draw[-] ($(chair.south)$) -- ($(patient2.north)+(0.04em,0)$);
      \draw[-] ($(moved.south)$) -- ($(agent1.north)$);
      \draw[-] ($(moved.south)$) -- ($(patient2.north)-(0,0.03em)$);
      \draw[-] ($(moved.south)$) -- ($(agent1.north)$);
      \draw[-] ($(moved.south)-(0,0.03em)$) -- ($(patient1.north)$);
      \draw[-] ($(moved.south)-(0,0.03em)$) -- ($(agent2.north)$);
      \draw[-] ($(neq.south)$) -- ($(patient1.north)+(0.08em,0)$);
      \draw[-] ($(neq.south)$) -- ($(patient2.north)+(0.08em,0)$);
    \end{tikzpicture}}
    \end{minipage}
  \end{tabular}\vspace{0.8ex}
  \caption{
    (left)~Tracker lattices for every sentence participant are combined with predicate HMMs.
    The MAP estimate in the resulting cross-product lattice
    simultaneously finds the best tracks and the best state sequences for every predicate.
    (right)~Two interpretations of the sentence ``Claire and Bill moved a chair'' having
    different first order logic formulas.
    The top interpretation corresponds to Bill and Claire moving the
    same chair, while the bottom one describes them moving different chairs.
    Predicates are highlighted in blue at the top and variables are highlighted in red at the bottom.
    Each predicate has a corresponding HMM which recognizes its presence in a video.
    Each variable has a corresponding tracker which locates it in a video.
    Lines connect predicates and the variables which fill their argument slots.
    Some predicates, such as \emph{move} and \emph{$\neq$}, take multiple arguments.
    Some predicates, such as \emph{move}, are applied multiple times between
    different pairs of variables.
  }
  \label{fig:model}
\end{figure*}

To perform the disambiguation task, we extend the sentence recognition
model of Siddharth \etal (2014) which represents sentences as
compositions of words.
Given a sentence, its first order logic interpretation and a video, our model produces a score which
determines if the sentence is depicted by the video.
It simultaneously tracks the participants in the events described by
the sentence while recognizing the events themselves.
This allows it to be flexible in the presence of noise by integrating
top-down information from the sentence with bottom-up information from
object and property detectors.
Each word in the query sentence is represented by an HMM \cite{BaumPSW70}, 
which recognizes tracks (i.e. paths of detections in a video for a specific object) 
that satisfy the semantics of the given word.
In essence, this model can be described as having two layers, one in
which object tracking occurs and one in which words observe tracks and
filter tracks that do not satisfy the word constraints.

Given a sentence interpretation, we construct a sentence-specific model
which recognizes if a video depicts the sentence as follows.
Each predicate in the first order logic formula has a corresponding HMM, 
which can recognize if that predicate is true of a video
given its arguments.
Each variable has a corresponding tracker which attempts to physically
locate the bounding box corresponding to that variable in each
frame of a video.
This creates a bipartite graph: HMMs that represent predicates are
connected to trackers that represent variables.
The trackers themselves are similar to the HMMs, in that they comprise a
lattice of potential bounding boxes in every frame.
To construct a joint model for a sentence interpretation, we take the cross product of HMMs
and trackers, taking only those cross products dictated by the
structure of the formula corresponding to the desired interpretation.
Given a video, we employ an object detector to generate candidate
detections in each frame, construct trackers which select one of these
detections in each frame, and finally construct the overall model from HMMs and
trackers.

Provided an interpretation and its corresponding formula composed of
$P$ predicates and $V$ variables, along with a collection of object
detections, $b^{\text{frame}}_{\text{detection index}}$, in each frame
of a video of length $T$ the model computes the score of the
video-sentence pair by finding the optimal detection for each
participant in every frame.
This is in essence the Viterbi algorithm \cite{Viterbi1971}, the MAP
algorithm for HMMs, applied to finding optimal object detections
$j^{\text{frame}}_{\text{variable}}$ for each participant, and the
optimal state $k^{\text{frame}}_{\text{predicate}}$ for each predicate HMM, in
every frame.
Each detection is scored by its confidence from the object detector,
$f$ and each object track is scored by a motion coherence metric $g$
which determines if the motion of the track agrees with the underlying
optical flow.
Each predicate, $p$, is scored by the probability of observing a particular
detection in a given state $h_p$, and by the probability of
transitioning between states $a_p$.
The structure of the formula and the fact that multiple predicates
often refer to the same variables is recorded by $\theta$, a mapping
between predicates and their arguments.
The model computes the MAP estimate as:
\begin{equation*}
\begin{small}
  \begin{aligned}
  \max_{\substack{j^1_1,\ldots,\;j^T_1\\\vdots\\j^1_V,\ldots,\;j^T_V}}
  \max_{\substack{k^1_1,\ldots,\;k^T_1\\\vdots\\k^1_P,\ldots,\;k^T_P}}
  \!\sum_{v=1}^V
  \!\sum_{t=1}^Tf(b^t_{j^t_v})+
  \!\sum_{t=2}^Tg(b^{t-1}_{j^{t-1}_v},b^t_{j^t_v})
  +\\
  \!\sum_{p=1}^P
  \!\sum_{t=1}^T
  h_p(k^t_p,b^t_{j^t_{\theta^1_p}},b^t_{j^t_{\theta^2_p}})+
  \!\sum_{t=2}^Ta_p(k^{t-1}_p,k^t_p)
  \end{aligned}
\end{small}
\end{equation*}
for sentences which have words that refer to at most two tracks
(i.e. transitive verbs or binary predicates) but is trivially extended
to arbitrary arities.
Figure~\ref{fig:model} provides a visual overview of the model as a
cross-product of tracker models and word models.

Our model extends the approach of Siddharth \etal (2014)
in several ways. First, we depart from the dependency based representation used in that work,
and recast the model to encode first order logic formulas. Note that some complex first order logic formulas cannot be directly
encoded in the model and require additional inference steps. 
This extension enables us to represent ambiguities in which a given sentence has multiple logical
interpretations for the same syntactic parse.

Second, we introduce several model components which are not specific 
to disambiguation, but are required to encode linguistic constructions that
are present in our corpus and could not be handled by the model of Siddharth \etal (2014). 
These new components are the predicate ``not equal'', disjunction, and
conjunction. The key addition among these components is support for the new predicate ``not equal'',
which enforces that two tracks, \ie objects, are distinct from each other.
For example, in the sentence ``Claire and Bill moved a chair'' one
would want to ensure that the two movers are distinct entities.
In earlier work, this was not required because the sentences tested in
that work were designed to distinguish objects based on constraints
rather than identity.
In other words, there might have been two different people but they
were distinguished in the sentence by their actions or appearance.
To faithfully recognize that two actors are moving the chair in the
earlier example, we must ensure that they are disjoint from each other.
In order to do this we create a new HMM for this predicate, which
assigns low probability to tracks that heavily overlap, forcing the
model to fit two different actors in the previous example.
By combining the new first order logic based semantic representation in lieu
of a syntactic representation with a more expressive model, we can encode the sentence 
interpretations required to perform the disambiguation task.

Figure~\ref{fig:model}(left) shows an example of two different
interpretations of the above discussed sentence ``Claire and Bill moved a chair''.
Object trackers, which correspond to variables in the first order
logic representation of the sentence interpretation, are shown
in red.
Predicates which constrain the possible bindings of the trackers,
corresponding to predicates in the representation of the sentence, are
shown in blue.
Links represent the argument structure of the first order logic formula,
and determine the cross products that are taken between the predicate HMMs
and tracker lattices in order to form the joint model which recognizes the entire
interpretation in a video.

The resulting model provides a single unified formalism for
representing all the ambiguities in table~\ref{tab:sentence-examples}.
Moreover, this approach can be tuned to different levels of
specificity.
We can create models that are specific to one interpretation of a
sentence or that are generic, and accept multiple interpretations by
eliding constraints that are not common between the different
interpretations.
This allows the model, like humans, to defer deciding on a particular
interpretation or to infer that multiple interpretation of the sentence
are plausible.

\section{Experimental Results}
\label{sec:results}


We tested the performance of the model described in the previous
section on the \emph{LAVA} dataset presented in section \ref{sec:corpus}.
Each video in the dataset was pre-processed with object detectors for
humans, bags, chairs, and telescopes.
We employed a mixture of CNN \cite{krizhevsky2012imagenet} and DPM
\cite{felzenszwalb2010object} detectors, trained on held out sections
of our corpus.
For each object class we generated proposals from both the CNN and the
DPM detectors, and trained a scoring function to map both results into
the same space.
The scoring function consisted of a sigmoid over the confidence of the
detectors trained on the same held out portion of the training set.
As none of the disambiguation examples discussed here rely on the
specific identity of the actors, we did not detect their identity. 
Instead, any sentence which contains names was automatically converted to one which contains
arbitrary ``person'' labels.

The sentences in our corpus have either two or three interpretations.
Each interpretation has one or more associated videos where the scene
was shot from a different angle, carried out either by different
actors, with different objects, or in different directions of motion.
For each sentence-video pair, we performed a 1-out-of-2 or 1-out-of-3 classification task
to determine which of the interpretations of the corresponding
sentence best fits that video.
Overall chance performance on our dataset is 49.04\%, slightly lower
than 50\% due to the 1-out-of-3 classification examples.

The model presented here achieved an accuracy of 75.36\%
over the entire corpus averaged across all error categories.
This demonstrates that the model is largely capable of capturing the
underlying task and that similar compositional cross-modal models may
do the same.
For each of the 3 major ambiguity classes we had an accuracy of
84.26\% for syntactic ambiguities, 72.28\% for semantic ambiguities,
and 64.44\% for discourse ambiguities.

The most significant source of model failures are poor object
detections.
Objects are often rotated and presented at angles that are 
difficult to recognize.
Certain object classes like the telescope are much more difficult to
recognize due to their small size and the fact that hands tend to
largely occlude them.
This accounts for the degraded performance of the semantic ambiguities
relative to the syntactic ambiguities, as many more semantic ambiguities
involved the telescope.
Object detector performance is similarly responsible for the lower
performance of the discourse ambiguities which relied much more on the
accuracy of the person detector as many sentences involve only people
interacting with each other without any additional objects.
This degrades performance by removing a helpful constraint for inference, 
according to which people tend to be close to the objects they are manipulating.
In addition, these sentences introduced more visual uncertainty as they
often involved three actors.

The remaining errors are due to the event models.
HMMs can fixate on short sequences of events which
seem as if they are part of an action, but in fact are just noise or the prefix
of another action.
Ideally, one would want an event model which has a global view of the
action, if an object went up from the beginning to the end of the
video while a person was holding it, it's likely that the object
was being picked up.
The event models used here cannot enforce this constraint, they merely
assert that the object was moving up for some number of frames; an
event which can happen due to noise in the object detectors.
Enforcing such local constraints instead of the global constraint of
the motion of the object over the video makes joint tracking and event
recognition tractable in the framework presented here but can lead to
errors.
Finding models which strike a better balance between local information
and global constraints while maintaining tractable inference remains
an area of future work.

\section{Conclusion}
\label{sec:conclusion}

We present a novel framework for studying ambiguous utterances expressed in a visual context. 
In particular, we formulate a new task for resolving structural ambiguities using visual signal.
This is a fundamental task for humans, involving complex cognitive processing, and is a key challenge 
for language acquisition during childhood.
We release a multimodal corpus that enables to address this task, as well as support further investigation of 
ambiguity related phenomena in visually grounded language processing. Finally, we present a unified approach for resolving 
ambiguous descriptions of videos, achieving good performance on our corpus.

While our current investigation focuses on structural \emph{inference}, we intend to extend this line of work 
to \emph{learning} scenarios, in which the agent has to deduce the meaning of words and sentences from 
structurally ambiguous input. Furthermore, our framework can be beneficial for image and video retrieval applications in which the query is expressed in natural language.
Given an ambiguous query, our approach will enable matching and clustering the retrieved results
according to the different query interpretations.

\section*{Acknowledgments}
This material is based upon work supported by the Center for Brains, Minds, and Machines (CBMM), 
funded by NSF STC award CCF-1231216. SU was also supported by ERC Advanced Grant 269627 Digital Baby.
\newpage
\bibliographystyle{acl}
\bibliography{emnlp2015}

\end{document}

%% file: viterbi-event.tex
\begin{picture}(0,0)%
\includegraphics{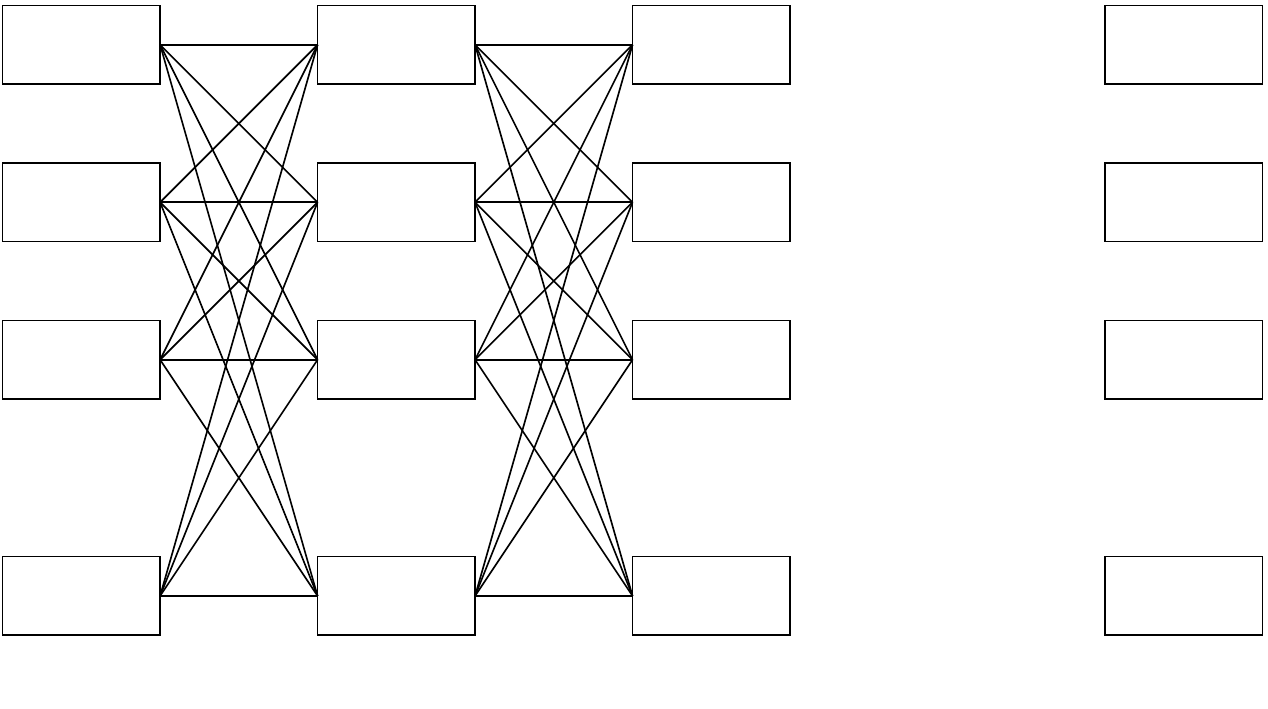}%
\end{picture}%
\setlength{\unitlength}{4144sp}%
\begingroup\makeatletter\ifx\SetFigFont\undefined%
\gdef\SetFigFont#1#2#3#4#5{%
  \reset@font\fontsize{#1}{#2pt}%
  \fontfamily{#3}\fontseries{#4}\fontshape{#5}%
  \selectfont}%
\fi\endgroup%
\begin{picture}(5784,3226)(979,-3365)
\put(1351,-2401){\makebox(0,0)[b]{\smash{{\SetFigFont{12}{14.4}{\familydefault}{\mddefault}{\updefault}{\color[rgb]{0,0,0}$\vdots$}%
}}}}
\put(2791,-2401){\makebox(0,0)[b]{\smash{{\SetFigFont{12}{14.4}{\familydefault}{\mddefault}{\updefault}{\color[rgb]{0,0,0}$\vdots$}%
}}}}
\put(4231,-2401){\makebox(0,0)[b]{\smash{{\SetFigFont{12}{14.4}{\familydefault}{\mddefault}{\updefault}{\color[rgb]{0,0,0}$\vdots$}%
}}}}
\put(6391,-2401){\makebox(0,0)[b]{\smash{{\SetFigFont{12}{14.4}{\familydefault}{\mddefault}{\updefault}{\color[rgb]{0,0,0}$\vdots$}%
}}}}
\put(5311,-421){\makebox(0,0)[b]{\smash{{\SetFigFont{12}{14.4}{\familydefault}{\mddefault}{\updefault}{\color[rgb]{0,0,0}$\ldots$}%
}}}}
\put(5311,-1141){\makebox(0,0)[b]{\smash{{\SetFigFont{12}{14.4}{\familydefault}{\mddefault}{\updefault}{\color[rgb]{0,0,0}$\ldots$}%
}}}}
\put(5311,-1861){\makebox(0,0)[b]{\smash{{\SetFigFont{12}{14.4}{\familydefault}{\mddefault}{\updefault}{\color[rgb]{0,0,0}$\ldots$}%
}}}}
\put(5311,-2941){\makebox(0,0)[b]{\smash{{\SetFigFont{12}{14.4}{\familydefault}{\mddefault}{\updefault}{\color[rgb]{0,0,0}$\ldots$}%
}}}}
\put(1351,-3301){\makebox(0,0)[b]{\smash{{\SetFigFont{12}{14.4}{\familydefault}{\mddefault}{\updefault}{\color[rgb]{0,0,0}$h$}%
}}}}
\put(2071,-3301){\makebox(0,0)[b]{\smash{{\SetFigFont{12}{14.4}{\familydefault}{\mddefault}{\updefault}{\color[rgb]{0,0,0}$a$}%
}}}}
\end{picture}%

%% file: viterbi-object.tex
\begin{picture}(0,0)%
\includegraphics{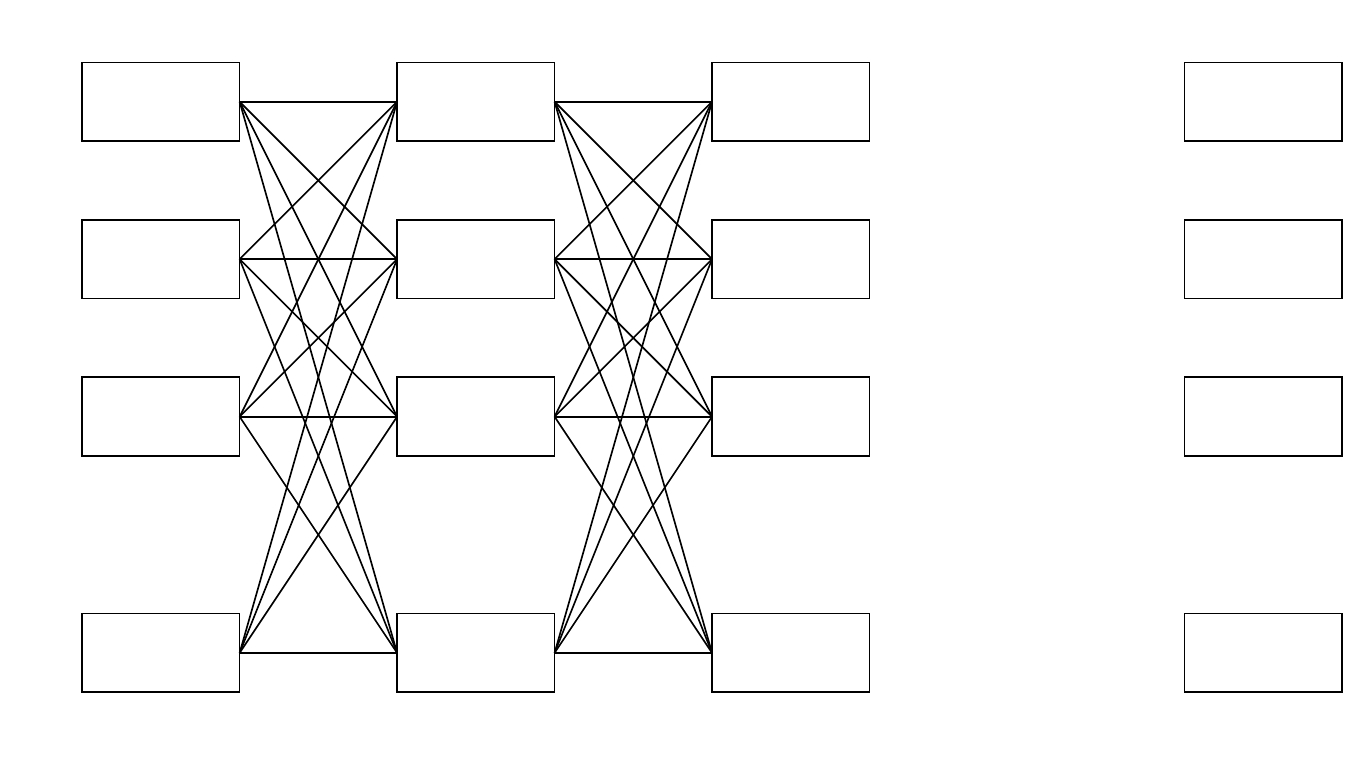}%
\end{picture}%
\setlength{\unitlength}{4144sp}%
\begingroup\makeatletter\ifx\SetFigFont\undefined%
\gdef\SetFigFont#1#2#3#4#5{%
  \reset@font\fontsize{#1}{#2pt}%
  \fontfamily{#3}\fontseries{#4}\fontshape{#5}%
  \selectfont}%
\fi\endgroup%
\begin{picture}(6147,3505)(616,-3374)
\put(1351,-2401){\makebox(0,0)[b]{\smash{{\SetFigFont{12}{14.4}{\familydefault}{\mddefault}{\updefault}{\color[rgb]{0,0,0}$\vdots$}%
}}}}
\put(2791,-2401){\makebox(0,0)[b]{\smash{{\SetFigFont{12}{14.4}{\familydefault}{\mddefault}{\updefault}{\color[rgb]{0,0,0}$\vdots$}%
}}}}
\put(4231,-2401){\makebox(0,0)[b]{\smash{{\SetFigFont{12}{14.4}{\familydefault}{\mddefault}{\updefault}{\color[rgb]{0,0,0}$\vdots$}%
}}}}
\put(6391,-2401){\makebox(0,0)[b]{\smash{{\SetFigFont{12}{14.4}{\familydefault}{\mddefault}{\updefault}{\color[rgb]{0,0,0}$\vdots$}%
}}}}
\put(5311,-421){\makebox(0,0)[b]{\smash{{\SetFigFont{12}{14.4}{\familydefault}{\mddefault}{\updefault}{\color[rgb]{0,0,0}$\ldots$}%
}}}}
\put(5311,-1141){\makebox(0,0)[b]{\smash{{\SetFigFont{12}{14.4}{\familydefault}{\mddefault}{\updefault}{\color[rgb]{0,0,0}$\ldots$}%
}}}}
\put(5311,-1861){\makebox(0,0)[b]{\smash{{\SetFigFont{12}{14.4}{\familydefault}{\mddefault}{\updefault}{\color[rgb]{0,0,0}$\ldots$}%
}}}}
\put(5311,-2941){\makebox(0,0)[b]{\smash{{\SetFigFont{12}{14.4}{\familydefault}{\mddefault}{\updefault}{\color[rgb]{0,0,0}$\ldots$}%
}}}}
\put(1351,-3301){\makebox(0,0)[b]{\smash{{\SetFigFont{12}{14.4}{\familydefault}{\mddefault}{\updefault}{\color[rgb]{0,0,0}$f$}%
}}}}
\put(2071,-3301){\makebox(0,0)[b]{\smash{{\SetFigFont{12}{14.4}{\familydefault}{\mddefault}{\updefault}{\color[rgb]{0,0,0}$g$}%
}}}}
\put(1351,-16){\makebox(0,0)[b]{\smash{{\SetFigFont{12}{14.4}{\rmdefault}{\mddefault}{\updefault}{\color[rgb]{0,0,0}$t=1$}%
}}}}
\put(2791,-16){\makebox(0,0)[b]{\smash{{\SetFigFont{12}{14.4}{\rmdefault}{\mddefault}{\updefault}{\color[rgb]{0,0,0}$t=2$}%
}}}}
\put(4231,-16){\makebox(0,0)[b]{\smash{{\SetFigFont{12}{14.4}{\rmdefault}{\mddefault}{\updefault}{\color[rgb]{0,0,0}$t=3$}%
}}}}
\put(6391,-16){\makebox(0,0)[b]{\smash{{\SetFigFont{12}{14.4}{\rmdefault}{\mddefault}{\updefault}{\color[rgb]{0,0,0}$t=T$}%
}}}}
\put(631,-376){\makebox(0,0)[b]{\smash{{\SetFigFont{12}{14.4}{\rmdefault}{\mddefault}{\updefault}{\color[rgb]{0,0,0}}%
}}}}
\put(631,-1816){\makebox(0,0)[b]{\smash{{\SetFigFont{12}{14.4}{\rmdefault}{\mddefault}{\updefault}{\color[rgb]{0,0,0}}%
}}}}
\put(631,-1096){\makebox(0,0)[b]{\smash{{\SetFigFont{12}{14.4}{\rmdefault}{\mddefault}{\updefault}{\color[rgb]{0,0,0}}%
}}}}
\put(631,-2896){\makebox(0,0)[b]{\smash{{\SetFigFont{12}{14.4}{\rmdefault}{\mddefault}{\updefault}{\color[rgb]{0,0,0}}%
}}}}
\put(1351,-376){\makebox(0,0)[b]{\smash{{\SetFigFont{12}{14.4}{\rmdefault}{\mddefault}{\updefault}{\color[rgb]{0,0,0}}%
}}}}
\put(1351,-1096){\makebox(0,0)[b]{\smash{{\SetFigFont{12}{14.4}{\rmdefault}{\mddefault}{\updefault}{\color[rgb]{0,0,0}}%
}}}}
\put(1351,-1816){\makebox(0,0)[b]{\smash{{\SetFigFont{12}{14.4}{\rmdefault}{\mddefault}{\updefault}{\color[rgb]{0,0,0}}%
}}}}
\put(1351,-2896){\makebox(0,0)[b]{\smash{{\SetFigFont{12}{14.4}{\rmdefault}{\mddefault}{\updefault}{\color[rgb]{0,0,0}}%
}}}}
\put(2791,-376){\makebox(0,0)[b]{\smash{{\SetFigFont{12}{14.4}{\rmdefault}{\mddefault}{\updefault}{\color[rgb]{0,0,0}}%
}}}}
\put(2791,-1096){\makebox(0,0)[b]{\smash{{\SetFigFont{12}{14.4}{\rmdefault}{\mddefault}{\updefault}{\color[rgb]{0,0,0}}%
}}}}
\put(2791,-1816){\makebox(0,0)[b]{\smash{{\SetFigFont{12}{14.4}{\rmdefault}{\mddefault}{\updefault}{\color[rgb]{0,0,0}}%
}}}}
\put(2791,-2896){\makebox(0,0)[b]{\smash{{\SetFigFont{12}{14.4}{\rmdefault}{\mddefault}{\updefault}{\color[rgb]{0,0,0}}%
}}}}
\put(4231,-376){\makebox(0,0)[b]{\smash{{\SetFigFont{12}{14.4}{\rmdefault}{\mddefault}{\updefault}{\color[rgb]{0,0,0}}%
}}}}
\put(4231,-1096){\makebox(0,0)[b]{\smash{{\SetFigFont{12}{14.4}{\rmdefault}{\mddefault}{\updefault}{\color[rgb]{0,0,0}}%
}}}}
\put(4231,-1816){\makebox(0,0)[b]{\smash{{\SetFigFont{12}{14.4}{\rmdefault}{\mddefault}{\updefault}{\color[rgb]{0,0,0}}%
}}}}
\put(4231,-2896){\makebox(0,0)[b]{\smash{{\SetFigFont{12}{14.4}{\rmdefault}{\mddefault}{\updefault}{\color[rgb]{0,0,0}}%
}}}}
\put(6391,-376){\makebox(0,0)[b]{\smash{{\SetFigFont{12}{14.4}{\rmdefault}{\mddefault}{\updefault}{\color[rgb]{0,0,0}}%
}}}}
\put(6391,-1096){\makebox(0,0)[b]{\smash{{\SetFigFont{12}{14.4}{\rmdefault}{\mddefault}{\updefault}{\color[rgb]{0,0,0}}%
}}}}
\put(6391,-1816){\makebox(0,0)[b]{\smash{{\SetFigFont{12}{14.4}{\rmdefault}{\mddefault}{\updefault}{\color[rgb]{0,0,0}}%
}}}}
\put(6391,-2896){\makebox(0,0)[b]{\smash{{\SetFigFont{12}{14.4}{\rmdefault}{\mddefault}{\updefault}{\color[rgb]{0,0,0}}%
}}}}
\end{picture}%